\documentclass[10pt, conference]{IEEEtran}
\usepackage{times}
\usepackage{graphicx}
\usepackage{subcaption}
\usepackage{amsmath}
\usepackage{amssymb}
\usepackage{amsfonts}
\usepackage[ruled, linesnumbered]{algorithm2e}
\usepackage{comment}
\usepackage{color}
\usepackage{multicol}
\usepackage{booktabs}
\usepackage[numbers]{natbib}
\usepackage{notoccite}
\usepackage[bookmarks=true]{hyperref}
\usepackage[font=small,labelfont=bf]{caption}

\usepackage[top=1in, left=0.75in, bottom=0.75in, right=0.75in]{geometry}
\IEEEoverridecommandlockouts
\begin{document}
\title{MRS-VPR: a multi-resolution sampling based global visual place recognition method}

\author{Peng Yin$^{1}$, Rangaprasad Arun Srivatsan$^3$, Yin Chen$^4$, Xueqian Li$^3$, \\ Hongda Zhang$^1$, Lingyun Xu$^{1}$, Lu Li$^3$, Zhenzhong Jia$^{3}$, Jianmin Ji$^{2,\mathbf{*}}$, Yuqing He$^{1,\mathbf{*}}$
\thanks{
This paper was supported by the National Natural Science Foundation of China (No. 61573386, No. 91748130, U1608253) and Guangdong Province Science and Technology Plan projects (No. 2017B010110011).

P. Yin, L. Xu, H. Zhang and Y. He are with the State Key Laboratory of Robotics, Shenyang Institute of Automation, Chinese Academy of Sciences, Shenyang, University of Chinese Academy of Sciences, Beijing. 
{(yinpeng, xulingyun, zhanghongda, heyuqing@sia.cn)}
J. Ji is with the School of Computer Science and Technology, University of Science and Technology of China, Hefei Anhui.
{(jianmin@ustc.edu.cn)}
R.A. Srivatsan, X. Li and L. Li are with the Biorobotics Lab, Robotics Institute, Carnegie Mellon University, Pittsburgh, PA 15213, USA.
{(arangapr, xueqianl, lilu12@andrew.cmu.edu)}
Y. Chen is with the School of Computer Science, University of Beijing University of Posts and Telecommunications, Beijing.
{(chenyin@bupt.edu.cn)}

{(Corresponding author: Jianmin Ji, Yuqing He)}
}
}

\maketitle


\begin{abstract}
Place recognition and loop closure detection are challenging for long-term visual navigation tasks. 
SeqSLAM is considered to be one of the most successful approaches to achieve long-term localization under varying environmental conditions and changing viewpoints. It depends on a brute-force, time-consuming sequential matching method.
We propose MRS-VPR, a multi-resolution, sampling-based place recognition method, which can significantly improve the matching efficiency and accuracy in sequential matching. 
The novelty of this method lies in the coarse-to-fine searching pipeline and a particle filter-based global sampling scheme, that can balance the matching efficiency and accuracy in the long-term navigation task. 
Moreover, our model works much better than SeqSLAM when the testing sequence has a much smaller scale than the reference sequence.
Our experiments demonstrate that the proposed method is efficient in locating short temporary trajectories within long-term reference ones without losing accuracy compared to SeqSLAM. 
\end{abstract}

\section{INTRODUCTION}
In mobile robotic systems, simultaneous localization and mapping (SLAM) is a process of constructing and updating the map of an unknown environment while performing localization~\cite{SLAM:VSLAM}. 
Visual place recognition (VPR) plays a vital role in finding reliable loop closures, helping SLAM to optimize the global localization and mapping. 
To improve the robustness against varying conditions, Milford \textit{et al.} proposed a sequence matching method, SeqSLAM~\cite{VPR:SeqSLAM}.
Given a set of $M$ reference frames and a set of $N$ testing frames, SeqSLAM can detect the potential matches based on feature similarities between the frames, with a computation complexity of $O(MN)$ using a brute-force searching method. 
As a result, sequential matching-based VPR is impractical in real robot navigation tasks because of two main challenges: (1) the exhaustive sequence searching method is computationally expensive with the stored frame sequence growing boundlessly, and (2) down-sampled frame sequence introduces uncertainty in matching process, when the length of testing frame sequence is too small.

\begin{figure*}[t]
	\centering
	\includegraphics[width=0.76\linewidth]{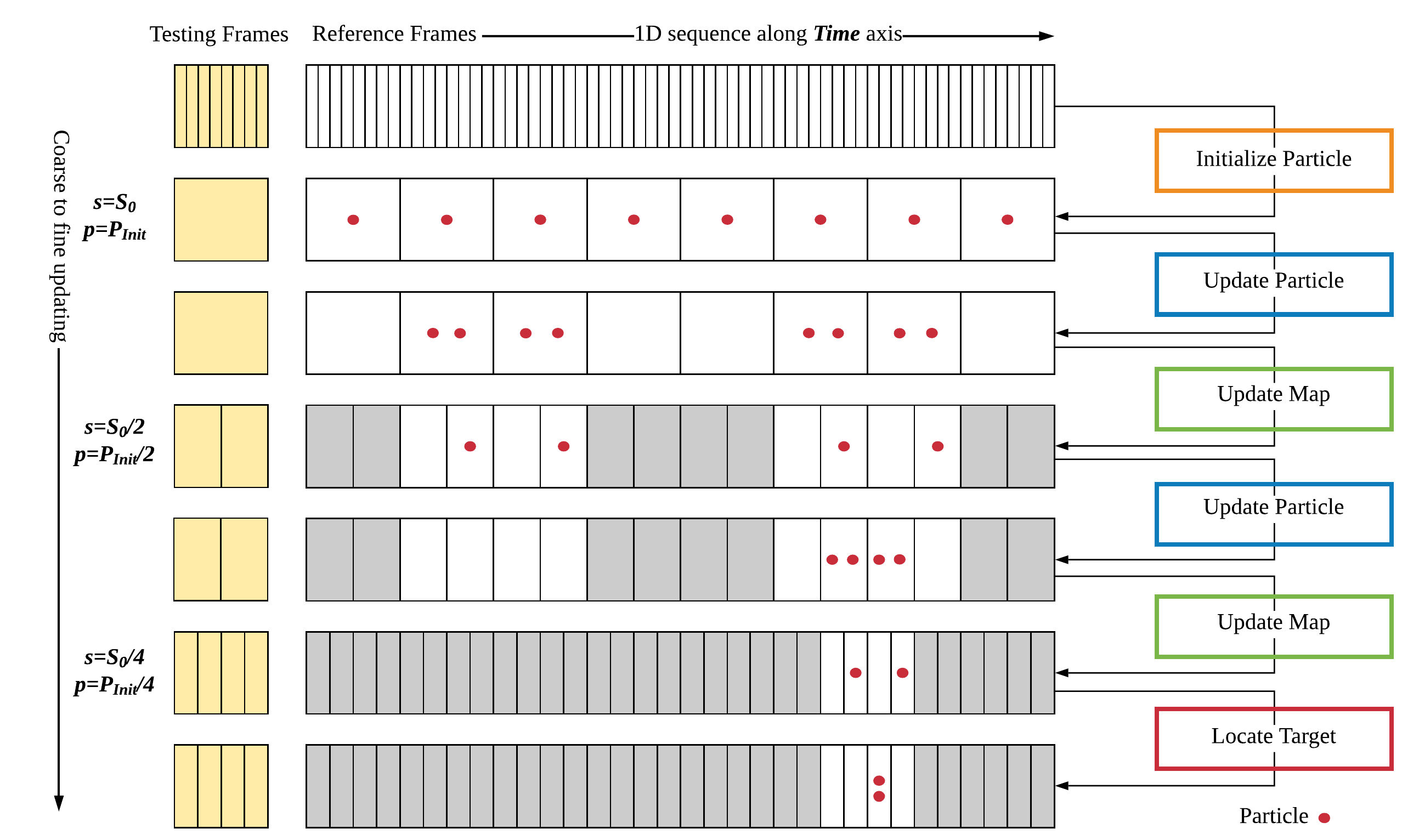}
	\caption{The Multi-Resolution Sampling method. 
		$s$ is the frame sampling interval, and $p$ is the number of particles.
		Row 1 shows the testing frames (yellow) and reference frames (white). 
		In row 2, both sequential frames are down-sampled to the lowest resolution level, and initial particles are uniformly sampled in the reference sequence, each particle represents a potential matching trajectory.
		By iterative updating the sequence resolution level, particles try to find the best match under highest resolution level.}
	\label{fig:framework}
\end{figure*}

To overcome these challenges, we apply a multi-resolution sampling (MRS) based method to improve the sequential matching efficiency.
In lower resolution level, the sequence matching of each particle can be evaluated quickly, resulting in fast convergence of the distribution of particles. 
This property helps the particles obtain good initial estimation at the beginning.
As see in Fig~\ref{fig:framework}, the computation complexity for each particle at the lower resolution level is smaller than at the higher resolution level. While the number of particles decreases at the denser resolution level, the overall efficiency of the dense frame sequence match is not compromised.
Therefore, we balance the matching efficiency and accuracy on higher resolution level.

The main contributions of this paper are listed as follows,
\begin{itemize}
	\item Based on the sequential matching-based place recognition method, we propose a multi-resolution sampling (MRS) scheme to balance the matching accuracy and searching efficiency.
	Our method is combined with a coarse-to-fine searching approach and a particle filtering scheme. This method is faster and more accurate than the original SeqSLAM method. It has wide potential for real world long-term robotic navigation tasks.
	\item We present a theoretical basis of our MRS-based method in the sequence matching. We also compare the improvement in performance of the MRS-VPR to the original SeqSLAM method.
	In the experiment part, we investigate the matching efficiency and accuracy of our proposed method, and discuss the key parameters within the MRS-VPR framework.
\end{itemize}

The rest of this paper is organized as follows. 
In Section II, we briefly introduce the recent developments in visual place recognition methods. 
In Section III, we describe our multi-resolution particle filter method for efficient sequence matching. 
Section IV demonstrates the performance of our method, details the experiment designs, and evaluates results, and conclusions are presented in Section V.

\section{Related Work}
\label{sec:related_work}




Appearance changing under variant conditions leads to an unstable place recognition in SLAM frameworks. Traditional V-SLAM methods use BoW~\cite{FeatureCapturer:BoW2} (vector of local handcrafted features) as the image descriptor,
or rely on prior 3D maps for online matching~\cite{pascoe2015farlap}, or use hierarchical BoW~\cite{garcia2017hierarchical}.

FABMAP~\cite{VPR:FABMAP} uses Bayesian filtering to achieve long-term place recognition over 1000 km~\cite{VPR:FABMAP}. However, FABMAP cannot handle scenarios with variant changes in environmental conditions. Another family of appearance-based place recognition method, SeqSLAM~\cite{VPR:SeqSLAM, naseer2014robust}, uses a series of frame sequence to improve the robustness under variant environments.
Lowry \textit{el al.}~\cite{VPR:Change_removal} assumed the differences caused by geometry features are relatively smaller than the differences caused by season-to-season appearance changes. They developed a Principal Component Analysis (PCA)~\cite{ML:PCA} approach to remove the season-related features, and extract remaining features as the season-invariant descriptions.

Sequential matching-based methods are not practical in real world applications due to their computational complexity.
To improve the robustness in sequence matching, Naseer~\textit{et al.}~\cite{naseer2014robust} proposed an minimum cost flow-based data association, which could deal with non-matching image sequences that result from temporal occlusions or from visiting new places. Vysotska~\textit{et al.} ~\cite{vysotska2015efficient}, improved the work of Naseer~\textit{et al.} with GPS priors.

Even though there is a rich literature that focus on dealing with varying conditional problems~\cite{VPR:Change_removal, naseer2014robust, vysotska2015efficient}, very few works focus on improving the efficiency and accuracy of searching in long-term place recognition tasks~\cite{VPR:Fast-SeqSLAM,VPR:PF_SeqSLAM}. 
Recently, with the development of deep learning for computer vision, Porav \textit{et al.}~\cite{VPR:Adversarial_VPR} improved feature robustness against variant conditions by extracting reliable convolution layer features.

More recently, Sayem~\cite{VPR:Fast-SeqSLAM} proposed a Fast-SeqSLAM method, which improved the searching efficiency by utilizing an approximate nearest neighbor (ANN) as the initial estimate for potential matches. 
Since ANN in Fast-SeqSLAM still relies on single image feature similarities, the initial search efficiency may decrease when the original matching frame sequence is of a relatively long-time scale.
Liu and Zhang~\cite{VPR:PF_SeqSLAM} applied a particle filter to improve the matching efficiency, where each particle represented a potential subset of the frame sequence~\cite{VPR:PF}.
Rather than evaluating the whole frame sequence, they predicted the weights of multiple particles based on frame sequence similarities and the robot motion.
However, both the methods described above require a good estimation of the initial matched location. 

\section{Proposed Method}
\label{sec:proposed_method}
Our work avoids the brute-force searching scheme in the traditional sequential matching methods by introducing a multi-resolution sampling approach, which combines a coarse-to-fine searching scheme and a particle filter method.
Each particle represents a potential frame sequence in reference frames.
As shown in Fig.~\ref{fig:framework} and Algorithm~\ref{Alg:MRS-SeqSLAM}, our method can be divided into following steps:
\begin{enumerate}
	\item Set the initial resolution level. down-sample trajectories according to the current resolution level and the initial particles; (line $5\sim6$)
    \item Update the particle status based on their evaluation results; (line $9\sim18$)
	\item Update the current resolution level and particle indexes. If the map resolution reaches the maximum level, go to step 4; else, go to step 2; (line $20\sim21$)
	\item Sort particles by their weightings, and predict the best particle. (line $24\sim25$)
\end{enumerate}

\begin{algorithm}[t]
	\SetKwData{Left}{left}\SetKwData{This}{this}\SetKwData{Up}{up}
	\SetKwFunction{Union}{Union}\SetKwFunction{FindCompress}{FindCompress}
	\SetKwInOut{Input}{Input}\SetKwInOut{Output}{Output}
	\Input{$M=$ Reference Frames, $N=$ Testing Frames}
	\Output{Predicted reference index}
	\Begin{
		$s=S_{0}$\;
		\For{$i\leftarrow 1$ \KwTo $l_{max}$}{
			\emph{/* Step1 Map Updating */}\;
			$m\longleftarrow$ M(s), $n\longleftarrow$ N(s), $s=s/2$\;
		    Generate initial particles $P_{init}$ according to Eq~\ref{eq:initial_paritcle}\; \label{Alg:MRS_init}
			\emph{/* Step2 Particle Updating */}\; 
			\While{ $M_{cover}>=50\%$ }{
				\ForEach{$p_{j}$ in $P$}{
				$t_{M}, t_{N} =$ Extract$(m, n, p_{j}.index)$\;
				$value, new\_index =$ Evaluate$(t_{M}, t_{N})$\; \label{Alg:MRS_evaluate}
				$p_{j}.weight = p_{j}.weight * value$\;	
				$p_{j}.index =$ new\_index\;
				}
				Particles weighting normalization according to Eq~\ref{eq:normalization}\;
				effectiveness = Evaluate particles efficiency according to Eq~\ref{eq:effect_particle}\;
				\If{effectiveness $<$ threshold\_effect}{
					Particles Resampling\;
				}
				Calculate map coverage according to Eq~\ref{eq:map_cover}\;
			}
			\emph{/* Step3 Map Updating */}\;
			Update sequence frames, and particles' status\;
		}
		Sort particles according to the particle weight\;
		\textbf{return} Best particle index
		\caption{MRS-VPR}\label{Alg:MRS-SeqSLAM}
	}
\end{algorithm}


\subsection{Particle Initialization}
In the particle initialization step, by setting the frame skipping interval as $s=S_{0}$, we can down-sample both the reference and the testing frame sequence into lowest map resolution level. 
At the lowest level, particles are sampled uniformly along the whole frame sequence.
Also, the initial number of particles $P_{init}$ satisfies the following equation,
\begin{align}
P_{init}=\frac{M}{N}\tau,\label{eq:initial_paritcle}
\end{align}
where $M$ and $N$ are the frame sequence length of reference frames and testing frames respectively; $\tau$ is the hyper parameter, which determines the overlaps between one potential neighbor frame sequence. 
The overlap rate of the neighbor frame sequence can be calculated by
\begin{align}
Overlaps=\frac{\tau-1}{\tau}. \label{eq:overlaps}
\end{align}
High overlaps will improve searching robustness, while reducing the matching time. 
In the experiment part, we will analyze the influence of the $\tau$ configuration in the place recognition task.

Initial particles are uniformly sampled from the reference frame sequence, as shown in the second row of Fig.~\ref{fig:framework}. Therefore, the entire particle sets have the following format, 
\begin{align}
	&P=\{p_{t}^{[1]},p_{t}^{[2]},p_{t}^{[3]},...,p_{t}^{[M]}\} \\ \nonumber
	&p_{t}^{i}=[index_{t}^{i}, weight_{t}^{i}],
\end{align}
where $index_{t}^{i}$ and $weight_{t}^{i}$ represent the end index and particle weight of the predicted reference frame sequence respectively. 

\subsection{Particle Evaluation}
\label{subsec:frame sequence_updating}
For each particle, we use the sequential matching to evaluate their measurements.
The goal for sequential matching is to estimate the frame sequence similarity with a given testing and local reference trajectories, and then update the predicted particle status.
As shown in Algorithm~\ref{Alg:Evaluation}, we use the pre-processed image data as the feature description. The feature distance is defined as the sum of absolute feature differences, similar to the original SeqSLAM~\cite{VPR:SeqSLAM}.
Although there are other feature extraction methods, this paper only focuses on the efficiency and accuracy in sequence matching. Thus, we only use the basic feature extraction method.

\begin{figure}
	\centering
	\includegraphics[width=0.6\linewidth]{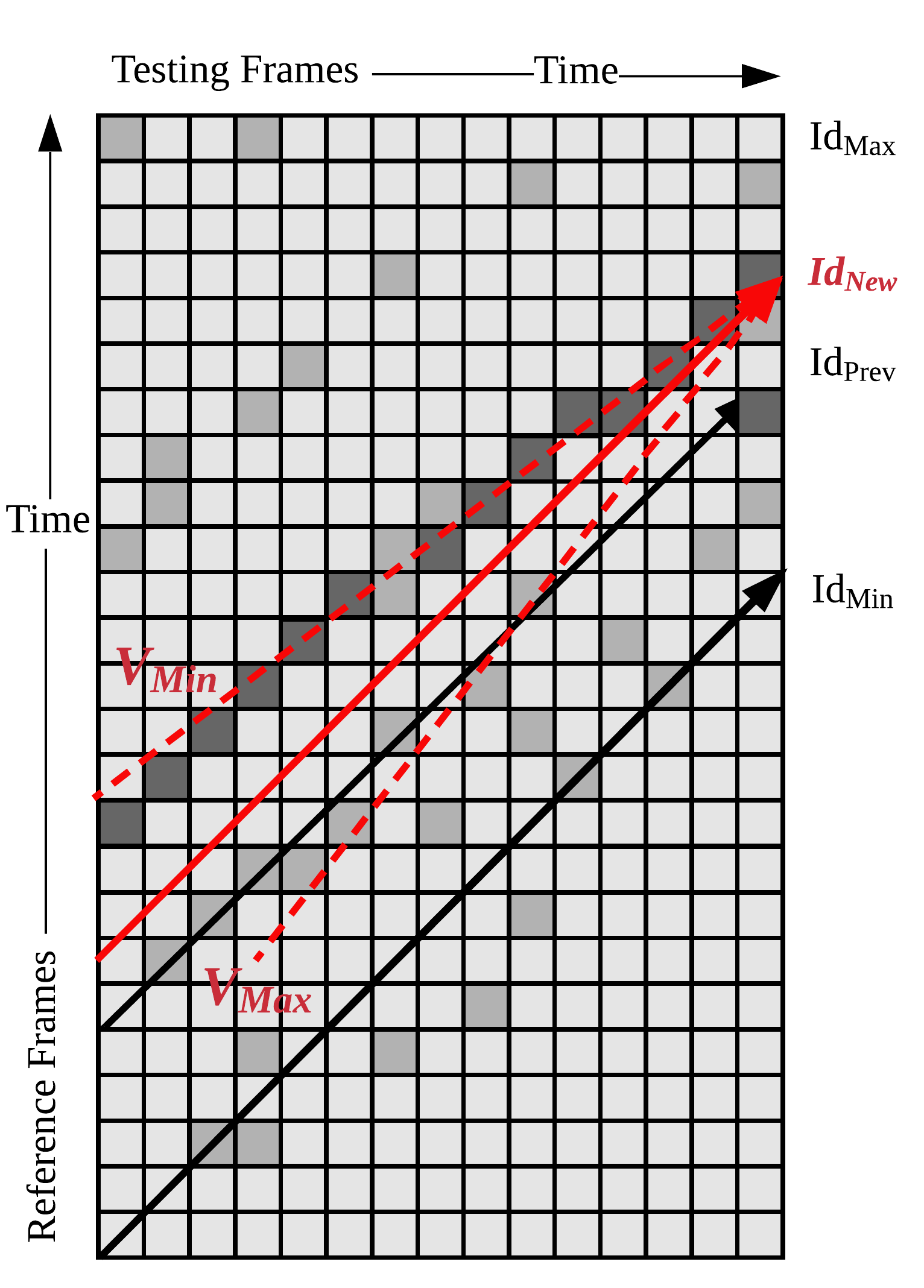}
	\caption{The sequence matching mechanism for the testing and local reference frames.}
	\label{fig:matching}
\end{figure}

With the extracted features, the difference matrix can be calculated based on the feature distance between reference and testing trajectories.
As shown in Fig.~\ref{fig:matching}, $x$-axis represents current testing frames $t_N$, and $y$-axis serves as potential reference frames $t_M$; 
the color of matrix cell represents the feature similarities, where darker colors imply relative place descriptions are more similar. $Id_{prev}$ is the index of previous reference prediction, $Id_{new}$ is the index of new predicted reference. 
Thus $Id_{new}$ is searched within $(Id_{prev}-Id_{shift}, Id_{prev}+Id_{shift})$.

To assign weights from the current frame sequence match, we retrieve different trajectories for each potential reference index. At each end index, we apply different speed proportional constants $\frac{V_{test}}{V_{ref}}\in \left( 0.8, 1.2\right)$ between testing and reference frames since the speed varies along the frame sequence. Thus the frame sequence similarity score can be evaluated by
\begin{align}
\label{eq:score}
	score(j, v) &= \sum_{t=1}^{L_{test}}D(t, j-v(L_{test}-t)), \\ \nonumber
	Id_{new} &= arg\min_{id} score,
\end{align}
where $D^{L_{test}\times[L_{test}+2\cdot Id_{shift}]}$ 
is the difference matrix, and $L_{test}$ is the length of testing frames. 
Finally, the new index of the particle is updated according to the smallest frame sequence difference.

\begin{algorithm}[t]
	\SetKwData{Left}{left}\SetKwData{This}{this}\SetKwData{Up}{up}
	\SetKwFunction{Union}{Union}\SetKwFunction{FindCompress}{FindCompress}
	\SetKwInOut{Input}{Input}\SetKwInOut{Output}{Output}
	\Input{$T_{M}=$ Reference Frames, $T_{N}=$ Testing Frames}
	\Output{Matching Value and refined index}
	\Begin{
		$d_{M}=$ descriptor($T_{M}$), $d_{N}=$ descriptor($T_{N}$)\;
		\BlankLine
		$D$ = GetDifferenceMatrix($d_{M}, d_{N}$)\;
		\BlankLine
		Values $= [\ ]$\;
		\ForEach{$j$ from $(Id_{prev}-Id_{shift})$ to $(Id_{prev}+Id_{shift})$}{
			\ForEach{$v$ in $[0.8, 0.9,..., 1.2]$}{
				value $=$ score$(j,v)$,\quad \text{From Eq.} \ \ref{eq:score} ;
			}
			Values.add($\min $ values)\;
		}
		best\_score = $\min $Values\;
		best\_index = $Id_{min}+ arg\min_{id}$ Values\;
		\BlankLine
		\textbf{return} best\_score, best\_index
		\caption{Evaluation}\label{Alg:Evaluation}
	}
\end{algorithm}

\subsection{Particle Filtering}
In the particle filtering step, for each particle (potential frame sequence), we use the sequence matching scheme in Sec.~\ref{subsec:frame sequence_updating} to evaluate the frame sequence score based on Eq.~\ref{eq:score}. 
Then the new particle weighting $\hat{\omega}_{k}^{i}$ is obtained by,
\begin{align}
\omega_{k}^{i}=\omega_{k-1}^{i} \times \frac{1}{1+e^{-score_{i}}},
\end{align}

After updating all particles, we normalize the weights of the particles by
\begin{align}
\omega_{k}^{i}=\frac{\hat{\omega}_{k}^{i}}{\sum \hat{\omega}_{k}^{i}},\label{eq:normalization}
\end{align}
and calculate the effectiveness of particles $\hat{N}_{eff}$ as
\begin{align}
\hat{N}_{eff}=\frac{1}{\sum \left(\omega_{k}^{i}\right)^2}. \label{eq:effect_particle}
\end{align}

On the same map resolution level, particles are re-sampled around effective particles when the particle efficiency $\hat{N}_{eff}$ is smaller than a given threshold $thres_{particle}$.
As shown in the third row of Fig.~\ref{fig:framework}, finally, the particles will converge to potential matching targets.

\subsection{Map Updating}
\label{subsubsec:Cutting_map}
Since computation complexity of sequence matching grows with the resolution level, 
we need to restrict particle sampling area to guarantee the matching efficiency.
After particles reach the stabilized status on current map level, we compute the map coverage according to valid particles.
We define a map coverage rate $M_{cover}$, which indicates the current particle convergence level
\begin{align}
 M_{cover} = \frac{M_{cur}}{M_{prev}},\label{eq:map_cover}
\end{align}
where $M_{cur}$ and $M_{prev}$ are current map coverage and previous map coverage separately.
If the convergence rate $M_{cover}$ is shrunk below a given threshold (in our experiments, this threshold is arbitrarily chosen to be $50\%$), 
we update both sequence frames into higher resolution level. 

\begin{figure*}
	\begin{minipage}{0.48\linewidth}
		\includegraphics[width=\linewidth]{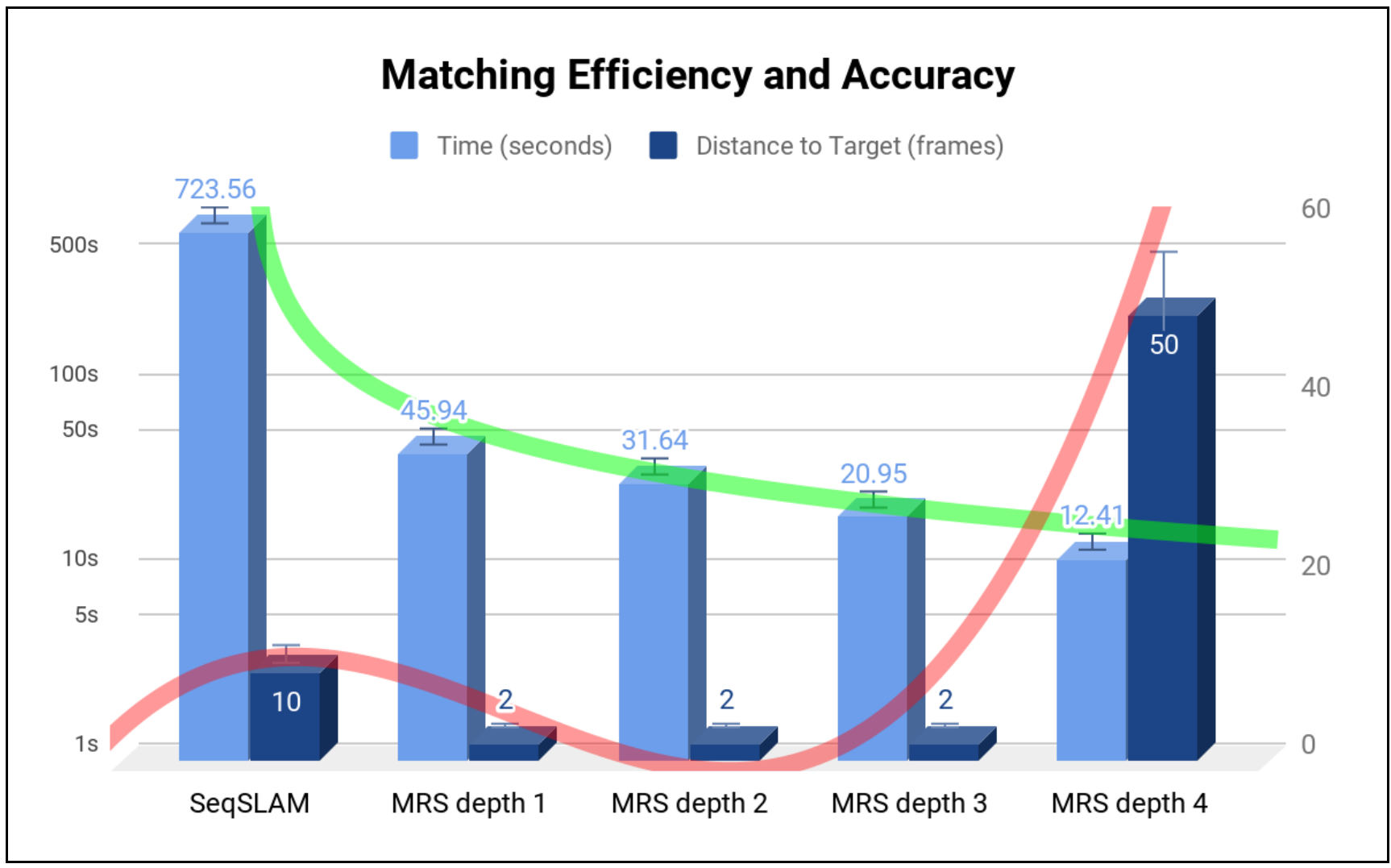}
		\caption{The efficiency and accuracy of place recognition results on the same \textit{Nordland} dataset~\cite{VPR:SeqSLAM_Robust}. 
		When the MAS map depth $l_{max}\leq 3$, the matching efficiency grows with the map resolution depth without reducing the matching efficiency.
		But when $l_{max}=4$, the matching error grows since testing frames are too sparse.}
		\label{fig:Results_AUC}
	\end{minipage}
	\hfill
	\begin{minipage}{0.48\linewidth}
		\includegraphics[width=\linewidth]{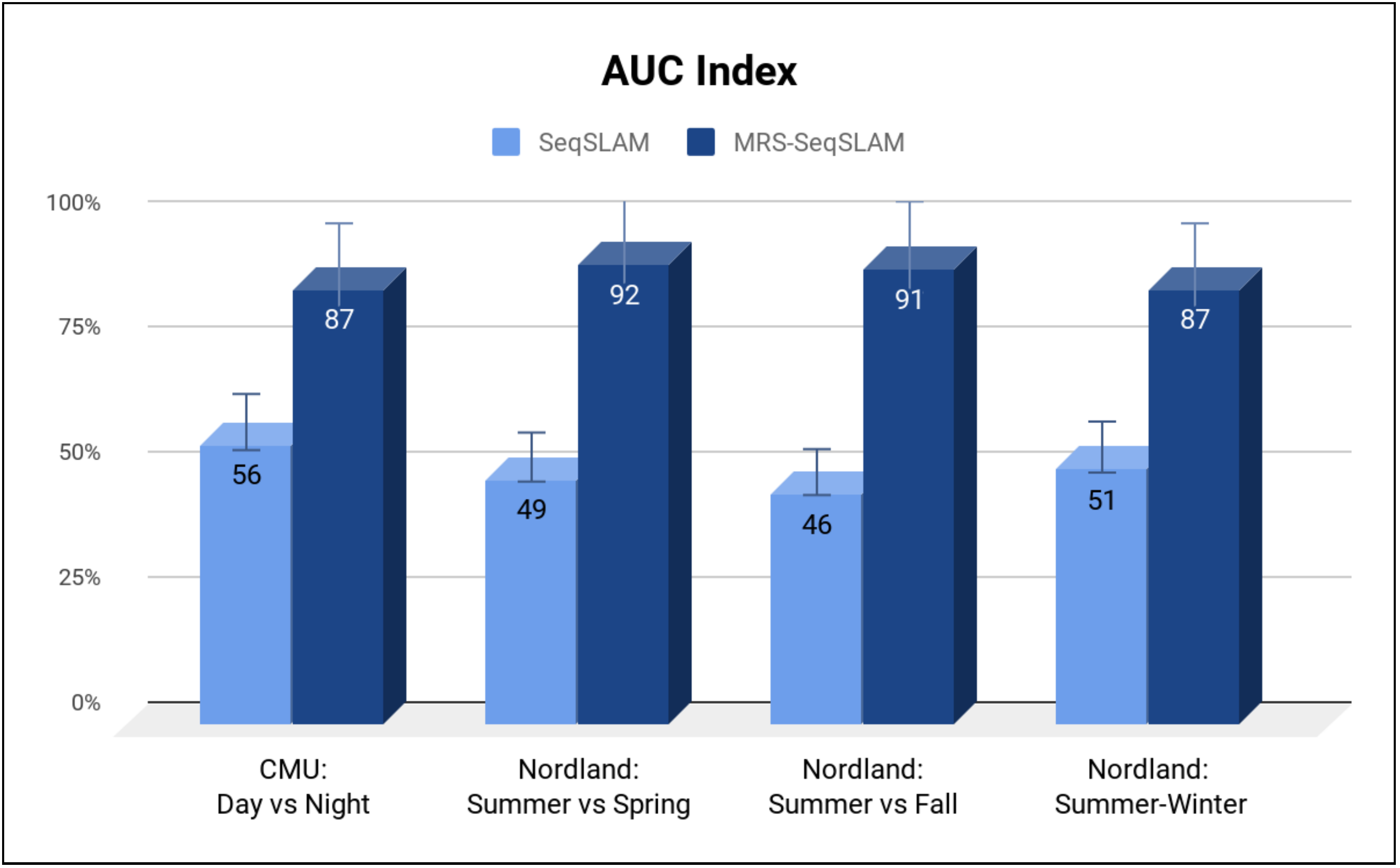}
		\caption{The accuracy of place recognition results. We use AUC (Area Under precision-recall curve) to indicate the matching performance. 
		For \textit{CMU} dataset, we test the place recognition performance under day-night condition. For \textit{Nordland} dataset，we test the matching results using summer versus spring, fall and winter conditions.}
        \label{fig:Results_Depth}
	\end{minipage}
\end{figure*}

\begin{figure*}[t]
	\begin{minipage}{0.45\linewidth}
		\includegraphics[width=\linewidth]{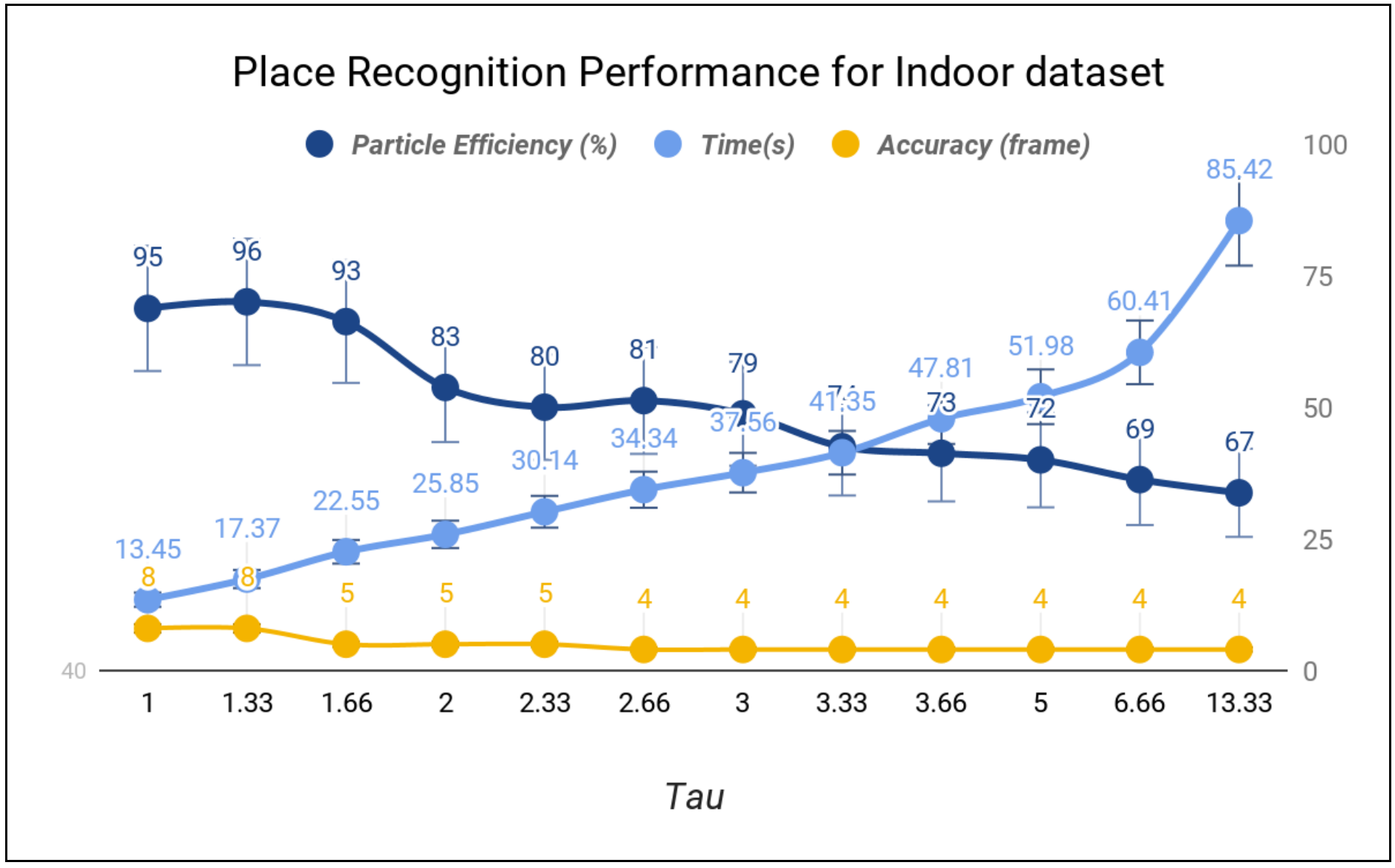}
	\end{minipage}
	\hfill
	\begin{minipage}{0.45\linewidth}
		\includegraphics[width=\linewidth]{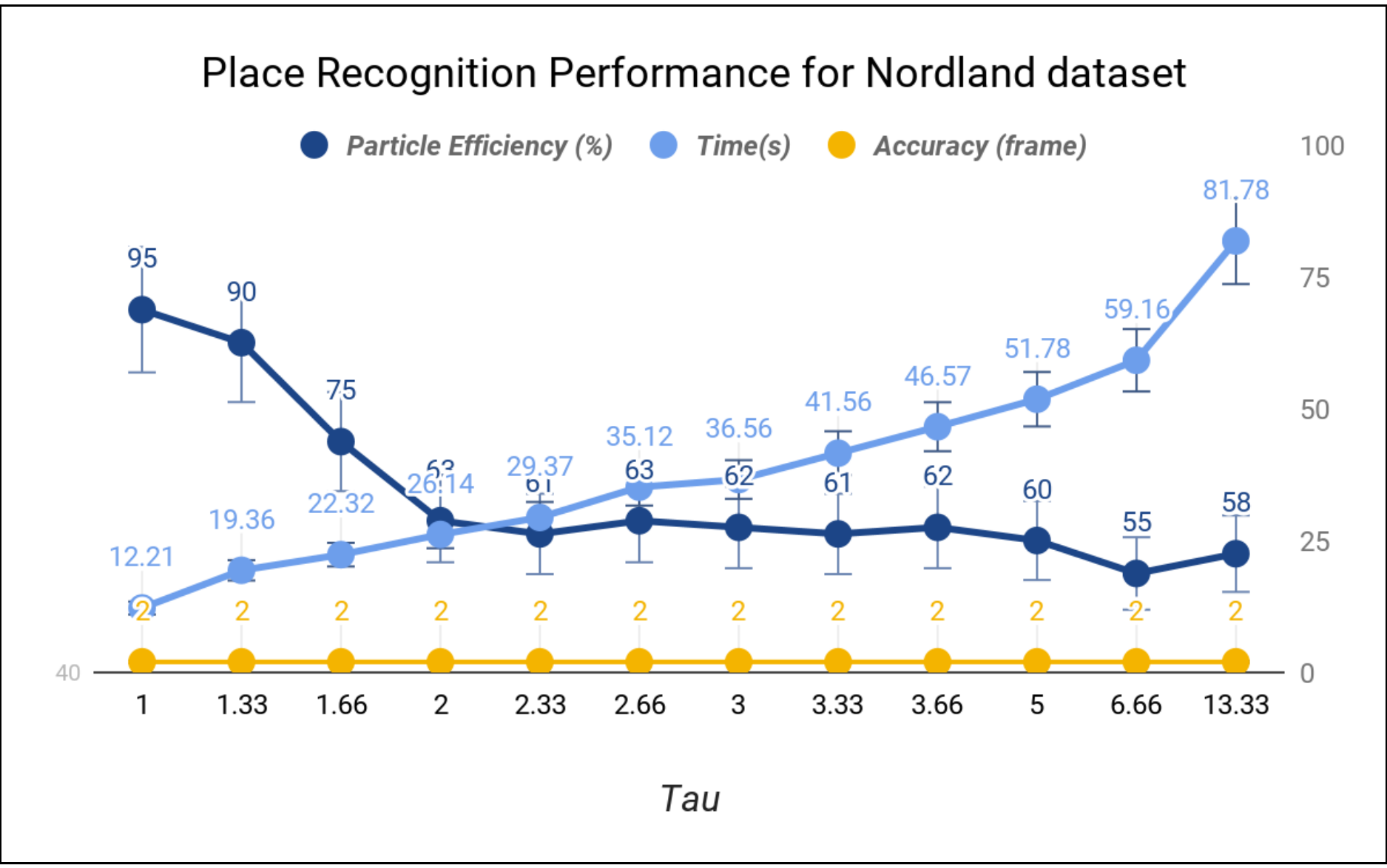}
	\end{minipage}
	\caption{The place recognition performance of the MRS-VPR method in \textit{CMU} and \textit{Nordland} dataset, with the same MRS depth level $l_{max}=3$.} 
	\label{fig:Init_particles}
\end{figure*}

\subsection{Speedup Analysis}
\label{subsec:Speedup}
In this subsection, we investigate the computation complexity of our MRS-VPR method by comparing it with the original SeqSLAM. 
For SeqSLAM, the computation complexity is $O\left(MN\right)$, where $N$ and $M$ are the  number of frames of testing and reference frame sequence, respectively.
For the MRS-VPR, we first generate $P_{init}$ initial particles on the whole reference sequence space.
Then for map resolution level $i$, the computation complexity is $O\left(\frac{P_{init}}{2^i}N_{i}\right)$, where $N_i$ is the testing frame index on the $i^{\text{th}}$ resolution level. $N_i=\frac{N}{2^{l_{max}-i}}$, while $l_{max}$ is the maximum resolution level.
The computation complexity ratio between the original \textit{SeqSLAM} method and our proposed \textit{MRS-VPR} is,
\begin{align*}
\mathcal{C}_{\frac{Seq}{MRS}}&=\frac{MN}{\sum_{i=0}^{l_{max}}\frac{P_{init}}{2^i}\cdot N_i}\label{eq:speedup} \\ \nonumber 
&= \frac{N}{\tau}\cdot \frac{2^{l_{max}}}{l_{max}},
\end{align*}
where we substitute for $P_{init}$ from Eq.~\ref{eq:initial_paritcle}.
For example, if we set $l_{max}=3$ and $\tau=2.0$, the computation complexity ratio will be $1.33N$.


\section{Experiments}
\label{sec:experiments}
In this section, we investigate the performance of our method in the long-term global place recognition task.
\subsection{Datasets}
We use two datasets to test our method: the \textit{Nordland} dataset~\cite{VPR:SeqSLAM_Robust}, which is a 728 km long train ride in northern Norway, covering the same route in four different seasons; a \textit{CMU} day-night dataset, which is a 1 km indoor sequence generated from a phone based camera with variant dynamics and viewpoint differences. We manually collected the second dataset, since we could not find large indoor VPR datasets containing day-night conditions.
Table.~\ref{table:dataset} shows the details about two datasets. 

\begin{table}[htbp]
	\caption[m1]{Datasets for place recognition task.}
	\renewcommand{\arraystretch}{1}
	\centering
	\begin{center}
		\begin{tabular}{ | c| c| c | c | c  |}
			\hline
			Dataset   & \textit{CMU}  & \textit{Nordland} \\ \hline
			Reference (frames)  & 9000 & 9000 \\ \hline
			Testing (frames)  & 300 & 300 \\ \hline
			Conditions & Day/Night & Four seasons \\ \hline
			Viewpoints & Not Fixed & Fixed	\\ \hline
 			Dynamics Objects & Yes & No	\\ \hline
		\end{tabular}
	\end{center}
	\label{table:dataset}
\end{table}
We observe that the main differences between \textit{Nordland} and \textit{CMU} dataset are viewpoints (fixed or not) and the existence of dynamics objects. 
While the \textit{Nordland} dataset is collected by mounting a camera in the cab of a train, the \textit{CMU} dataset is collected with a hand-held mobile phone camera. With the \textit{CMU} dataset, it is hard to guarantee a stable viewpoint along the route. In addition, \textit{CMU} dataset has lots of dynamic objects in the indoor environment. Thus, it is even harder to find potential matches in \textit{CMU} dataset, compared to the \textit{Nordland} dataset.

\subsection{Accuracy \& Efficiency Analysis}
To inspect the accuracy and efficiency of different algorithms, we leverage short-term testing frame sequence to match the relative long-term reference frame sequence.
For both \textit{CMU} and \textit{Nordland} datasets, the proportion between reference and testing frames number is 30, as shown in Table~\ref{table:dataset}.

One important parameter in our method is the map depth. Fig~\ref{fig:Results_Depth} shows the matching efficiency and accuracy of different methods in the \textit{CMU} datasets. 
We see that, when the map resolution depth $l_{max}=\{1,2,3\}$, the matching efficiency tends to improve with the growth of the map resolution depth, but the matching error is held at $2$ frames. In contrast, the original sequential matching method took $723.56s$, and the final error between predicted match and the best alignment matches is $10$ frames.
However, when $l_{max}=4$, the testing frames at the lowest level only have $\frac{300}{2^{4}}\sim 19$ frames. If the testing sequence is too small, the robustness of sequential matching against changing viewpoints and illuminations is lost;
and particles at the lowest resolution level cannot provide good estimations at the beginning. We also repeated the experiment on the \textit{Nordland} datasets. It turns out that $l_{max}=3$ is a suitable map depth for both the datasets.

Another important parameter in our method is $\tau$, which determines the initial number of particles. We investigate the matching performance under variant $\tau$ settings. 
As observed in Fig~\ref{fig:Init_particles}, with the increasing of $\tau$, the particle effectiveness index $\hat{N}_{eff}$ decreases. This means that there will be more particles converging to the potential optimal index. 
But when $\tau$ is less than $1.0$, the overlaps between two particles reduces to $0$, and the particles are less likely to converge to optimal positions.
In addition, the matching time also increases with $\tau$. 
In order to balance both efficiency and accuracy, we set $\tau$ within $(1.5, 2.5)$, depending on the requirement of efficiency. 
In our experiment, the default $\tau$ value is $2.0$.


Fig~\ref{fig:Results_AUC} shows the area under curve (AUC) index; higher the AUC index indicate more accurate matching results.
Compared to traditional sequential matching method, our method is more stable under varying conditions; all the AUC indexes are above $80\%$. 
This indicates that the coarse-to-fine searching scheme can improve the initial estimation for the best sequence matching. 
In summary, our method has the potential to balance the efficiency and accuracy in the long-term place recognition under variant environmental conditions.

\section{Conclusions}
\label{sec:conclusions}
In this paper, we propose a multi-resolution particle filter-based sequence matching method. Our framework leverages coarse-to-fine searching methods to improve the robustness of place recognition, when the testing sequence is much smaller than the reference sequence (e.g. overlap ratio $\frac{M}{N} > 30$). 
The experiments on \textit{Nordland} and \textit{CMU} datasets show that our MRS-VPR framework outperforms the appearance based sequence matching method SeqSLAM in the long-term place recognition task. 
For the future work, we plan to combine the proposed place recognition method with topological maps to construct more robust reference maps for real robot long-term navigation task.

\bibliographystyle{ieeetr}
\bibliography{main}

\begin{thebibliography}{10}

\bibitem{SLAM:VSLAM}
A.~J. Davison, ``Real-time simultaneous localisation and mapping with a single
  camera,'' in {\em Proceedings Ninth IEEE International Conference on Computer
  Vision}, pp.~1403--1410 vol.2, Oct 2003.

\bibitem{VPR:SeqSLAM}
M.~J. Milford and G.~F. Wyeth, ``{SeqSLAM: Visual route-based navigation for
  sunny summer days and stormy winter nights},'' in {\em IEEE International
  Conference on Robotics and Automation}, pp.~1643--1649, May 2012.

\bibitem{FeatureCapturer:BoW2}
H.~Jégou, F.~Perronnin, M.~Douze, J.~Sánchez, P.~Pérez, and C.~Schmid,
  ``Aggregating local image descriptors into compact codes,'' {\em IEEE
  Transactions on Pattern Analysis and Machine Intelligence}, vol.~34,
  pp.~1704--1716, Sept 2012.

\bibitem{pascoe2015farlap}
G.~Pascoe, W.~Maddern, A.~D. Stewart, and P.~Newman, ``{Farlap: Fast robust
  localisation using appearance priors},'' in {\em IEEE International
  Conference on Robotics and Automation (ICRA)}, pp.~6366--6373, IEEE, 2015.

\bibitem{garcia2017hierarchical}
E.~Garcia-Fidalgo and A.~Ortiz, ``Hierarchical place recognition for
  topological mapping,'' {\em IEEE Transactions on Robotics}, vol.~33, no.~5,
  pp.~1061--1074, 2017.

\bibitem{VPR:FABMAP}
M.~Nowakowski, C.~Joly, S.~Dalibard, N.~Garcia, and F.~Moutarde, ``{Topological
  localization using Wi-Fi and vision merged into FABMAP framework},'' in {\em
  IEEE/RSJ International Conference on Intelligent Robots and Systems (IROS)},
  pp.~3339--3344, Sept 2017.

\bibitem{naseer2014robust}
T.~Naseer, L.~Spinello, W.~Burgard, and C.~Stachniss, ``Robust visual robot
  localization across seasons using network flows,'' in {\em Twenty-Eighth AAAI
  Conference on Artificial Intelligence}, 2014.

\bibitem{VPR:Change_removal}
S.~Lowry and M.~Milford, ``Change removal: Robust online learning for changing
  appearance and changing viewpoint,'' {\em ICRA15 WS VPRiCE}, 2015.

\bibitem{ML:PCA}
S.~Wold, K.~Esbensen, and P.~Geladi, ``Principal component analysis,'' {\em
  Chemometrics and intelligent laboratory systems}, vol.~2, no.~1-3,
  pp.~37--52, 1987.

\bibitem{vysotska2015efficient}
O.~Vysotska, T.~Naseer, L.~Spinello, W.~Burgard, and C.~Stachniss, ``{Efficient
  and effective matching of image sequences under substantial appearance
  changes exploiting GPS priors},'' in {\em IEEE International Conference on
  Robotics and Automation (ICRA)}, pp.~2774--2779, IEEE, 2015.

\bibitem{VPR:Fast-SeqSLAM}
S.~M. Siam and H.~Zhang, ``{Fast-SeqSLAM: A fast appearance based place
  recognition algorithm},'' in {\em IEEE International Conference on Robotics
  and Automation (ICRA)}, pp.~5702--5708, May 2017.

\bibitem{VPR:PF_SeqSLAM}
Y.~Liu and H.~Zhang, ``{Towards improving the efficiency of sequence-based
  SLAM},'' in {\em IEEE International Conference on Mechatronics and
  Automation}, pp.~1261--1266, Aug 2013.

\bibitem{VPR:Adversarial_VPR}
H.~Porav, W.~Maddern, and P.~Newman, ``Adversarial training for adverse
  conditions: Robust metric localisation using appearance transfer,'' {\em
  arXiv preprint arXiv:1803.03341}, 2018.

\bibitem{VPR:PF}
R.~Van Der~Merwe, A.~Doucet, N.~De~Freitas, and E.~A. Wan, ``The unscented
  particle filter,'' in {\em Advances in neural information processing
  systems}, pp.~584--590, 2001.

\bibitem{VPR:SeqSLAM_Robust}
N.~S{\"u}nderhauf, P.~Neubert, and P.~Protzel, ``{Are we there yet? Challenging
  SeqSLAM on a 3000 km journey across all four seasons},'' in {\em Proc. of
  Workshop on Long-Term Autonomy, ICRA}, Citeseer, 2013.

\end{thebibliography}



\end{document}